\documentclass[conference]{IEEEtran}
\IEEEoverridecommandlockouts
\usepackage{cite}
\usepackage{amsmath,amssymb,amsfonts}
\usepackage{algorithmic}
\usepackage{graphicx}
\usepackage{textcomp}
\usepackage{xcolor}
\usepackage{booktabs}             
\usepackage{multirow}					

\def\BibTeX{{\rm B\kern-.05em{\sc i\kern-.025em b}\kern-.08em
    T\kern-.1667em\lower.7ex\hbox{E}\kern-.125emX}}
\begin{document}

\title{Visual Reasoning of Feature Attribution with Deep Recurrent Neural Networks
}

\author{Chuan Wang$^{1,2}$, Takeshi Onishi$^{1}$, Keiichi Nemoto$^{1}$ and Kwan-Liu Ma$^{2}$ \\
Fuji Xerox Co., Ltd$^{1}$  \hspace{0.5cm} University of California, Davis$^{2}$}


\maketitle

\begin{abstract}
Deep Recurrent Neural Network (RNN) has gained popularity in many sequence classification tasks. Beyond predicting a correct class for each data instance, data scientists also want to understand what differentiating factors in the data have contributed to the classification during the learning process. 
We present a visual analytics approach to facilitate this task by revealing the RNN attention for all data instances, their temporal positions in the sequences, and the attribution of variables at each value level. 
We demonstrate with real-world datasets that our approach can help data scientists to understand such dynamics in deep RNNs from the training results, hence guiding their modeling process.
\end{abstract}


\section{Introduction}
Sequence classification is a fundamental problem in Big Data analysis. 
Recent advances in Recurrent Neural Networks (RNNs) have achieved convincing results for such classification tasks. During the training process, RNN models capture the discriminating patterns that distinguishing them from different categories. 
It is thus increasingly used in many real-world domains such as funnel optimization for digital marketing, click-stream analysis for online purchase prediction, and patient treatment data analysis for medical recovery predictions.

\begin{figure}[tb]
 \centering 
 \includegraphics[width=\columnwidth]{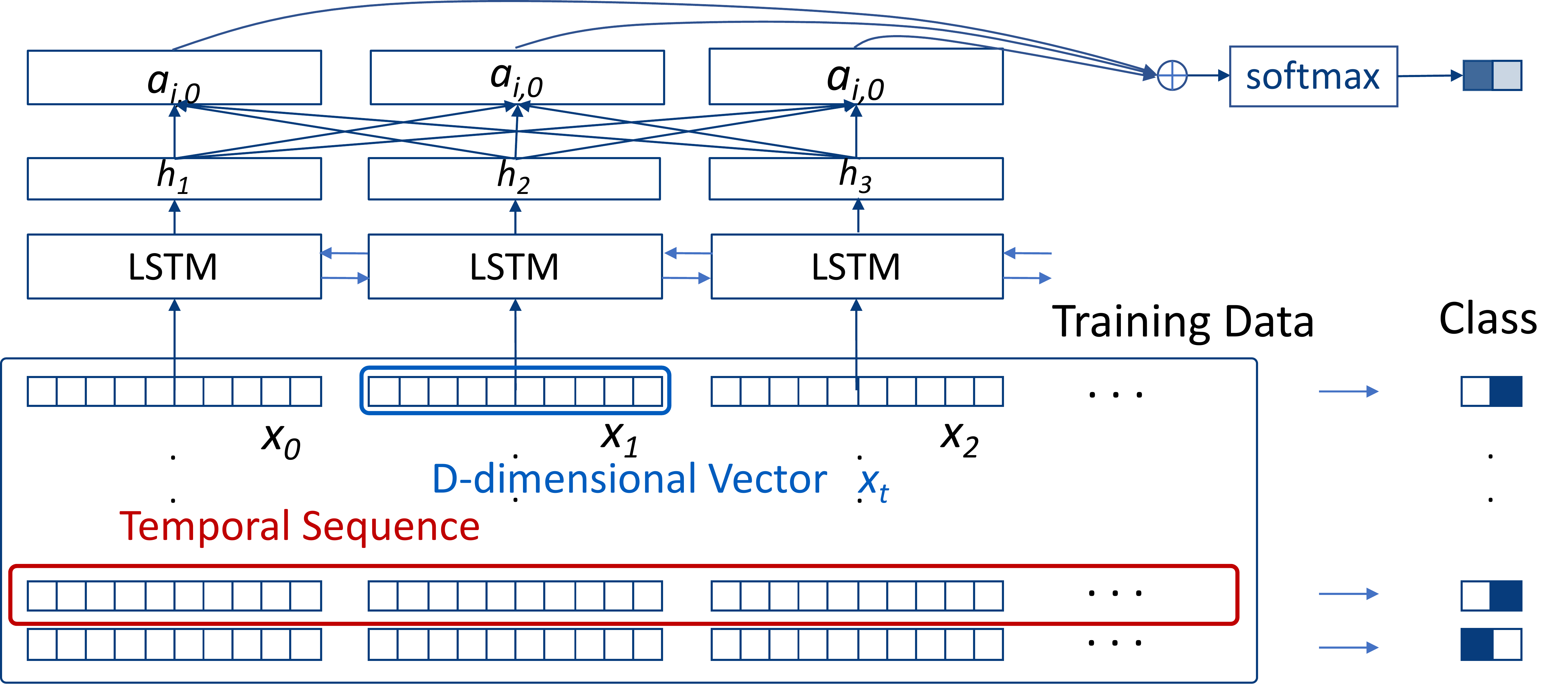}
 \vspace{-0.3cm}
 \caption{Training data instances: sequences of one-hot vectors and associated classes. Each dimension in the one-hot vectors represents a feature type.}
 \label{fig:training_data}
 \vspace{-0.7cm}
\end{figure}

Despite the popularity of RNN techniques, it can be challenging to understand what and how features are interpreted in learning with a high-performing model. As shown in Figure~\ref{fig:training_data}, one can encode multidimensional features as one-hot vectors, with categorical or numerical values vary at each element along the temporal (horizontal) axis and the data instance (vertical) axis. 
It is important to understand $which$ features contribute more and $how$ they contribute to the learning of RNN models in different scenarios. In practice, analysts have difficulties in feature selection.
Knowing which features contribute to high-performance helps analysts to select meaningful attributes during training. Also, the reasoning of important features can provide guidelines to business goals. For example, in sales performance management, analysts want to understand what kind of behavior, such as visiting or emailing, can help improve sales performance\cite{Wang2017}.

In this work, we focus on the visual reasoning of feature attribution for RNN models, motivated by the practical need in customer churn prediction and prevention. Analysts try to understand what customer service behaviors and their associated factors may lead to a high risk of churning for maintaining good customer relationship and increase customer loyalty.

Distilled from our interactions with the analysts, here are three challenges when building a visual analytics solution.

{\setlength{\parindent}{0cm}\textit{\textbf{Pattern Discovery from Multivariate Temporal Sequences.}} \enspace 
The training data is both complex in the feature dimensionality and in their temporal variance. 
Comparing the importance of different attributes helps in determining $what$ aspects to address for achieving a certain analytics goal, and clarifying their value level contributions explains $how$.
Analysts try to discover the effect of each attribute as well as the combined attributes. 
The value levels of these attributes change over time. Because RNNs are known to be able to learn the temporal patterns, the discovery of the contribution of temporal sequences becomes essential. }

{\setlength{\parindent}{0cm}\textit{\textbf{Mixture of Attribute Types.}} \enspace
Real-world dataset often contains three attribute types: numerical, categorical and ordinal. From the visual encoding perspective, it is challenging to unify the design for comparing across multiple types.}

{\setlength{\parindent}{0cm}\textit{\textbf{Multidimensional sequence.}} \enspace
Sequence data are often associated with metadata describing the \textit{result} or \textit{consequence} of the change in attributes over time. Merely visualizing the sequences themselves is insufficient in revealing the patterns.}

We present $AttributionHeatmap$ to address the challenges and surface the attribution of training data to the classification kernel modeling. specifically, we provide:

\begin{itemize}
\setlength\itemsep{0em}
\item \textbf{the design and implementation of a visual analytics system} that helps domain experts to understand the feature attribution in prediction tasks based on RNNs.
\item \textbf{case studies validating our hybrid design} which combines a 
multimatrix view and a multipartite graph view to reveal salient features, their contributing values, and temporal patterns.
\end{itemize}

\section{Related Work}

\textbf{Visualization for Deep Neural Networks (DNN)}
Many visualization techniques have been developed to facilitate the DNN model building process, covering domains such as image understanding \cite{TzengM05, ZeilerF20134268} and natural language processing \cite{Li2016VisualizingAU}. 
Techniques such as hierarchical correlation matrices \cite{Alsallakh2017}, edge-bundled DAG \cite{Liu2017}, parallel coordinates \cite{StrobeltGHPR16} and co-clustering \cite{Ming2017} were introduced for interpreting model specifics. Systems like ActiVis\cite{KahngAKC17} and \cite{Pezzotti2018} reveal the links between filters and patterns at data instance level.
On the RNN side, many literatures focus on model's attention mechanism to evaluate how well the inputs are related to the outputs. Xu et. al. \cite{pmlr-v37-xuc15} introduced an attention-based model and visualized the caption-word corresponding saliency over images. 
Bahdanau et. al. \cite{BahdanauCB14} visualized heatmap-like word attentions extracted by RNN models for natural language analysis. 
Yang et. al. \cite{Yang2016} visualize word-level and sentence-level attentions over texts.
Our work falls into this category in the spirit of analyzing attentions from RNNs and visualizes the saliency of data instances. Besides, our approach is capable of visualizing patterns from the training dataset instead of a few data instances to derive meaningful conclusions.

\textbf{Multivariate Data Visualization}
Multivariate data visualization have been developed in numerous fields of analysis\cite{Liu2017}. 
We summarize the related work based on the visualization layout.
\textit{Grid-based methods} arrange variables in matrices to benefit the pairwise comparison of attribute relationships. 
GPLOM \cite{Mcguffin201339} extends Scatterplot Matrix \cite{Wilkinson200686} to generalized plot matrices.
EnsembleMatrix \cite{Justin2009144} enables the interaction in matrices that help understand the classifiers in ensemble learning.
Yuan et. al. \cite{Yuan2009143} insert multidimensional scaling plot into neighboring parallel coordinate axes.
\textit{Glyph-based methods} encode multidimensional data with value-interpolated geometries \cite{Chan2013}.
DICON \cite{Cao2011} uses a treemap-like icon to encode data cluster that depicts multiple attributes and quality of cluster.
Irimia et. al. \cite{Irimia2012} adopted connectograms to visualize relationships between multidimensional neuron connectivities.
Since matrix and parallel coordinate plots (PCP) based methods are more scalable to larger datasets, we investigate both and further discuss the trade-offs in later chapters.

\textbf{Temporal Sequence Visualization}
We summarize temporal sequence visualizations into the following categories:
\textit{Juxtaposed representation} features in the visualization of events transfer. 
Alluvial diagrams \cite{Rosvall2010} reveal how network structures change over time.
Outflow~\cite{Wongsuphasawat2012} visualizes temporal event sequences in pathways that are similar to parallel coordinates. 
\textit{Matrix representation} emphasizes links between events. For example,  MatrixWave~\cite{Zhao2015} aligns and compares the differences in the occurrence of clickstreams augmented by event states. 
Liu et. al. ~\cite{Liu2016} presented an analytic pipeline for pattern mining, pattern pruning, and coordinated exploration between patterns and sequences. 
ViDX~\cite{Xu2016} extends Marey's graph with a time-aware outlier-preserving design to support faults detection and troubleshooting on assembly lines. 
The temporal sequence design in our work attempts to visualization and compare the temporal patterns of all data instances from multiple predicted classes. We also adopt a juxtaposed design to depict the temporal changes within the user-specified range.

\section{Background and Modeling}
\subsection{Customer Churn Prediction}

Predicting customer's likelihood of canceling a subscription to a service or a product is among the most studied Big Data problems. Data scientists build binary classification models to predict churning, in which feature engineering is key in identifying what may impact the churning behavior.

\begin{table}[tb]
  \caption{Attributes used in Customer Churn Prediction}
  \label{tab:CCP_attributes}
  \scriptsize%
  \centering%
  \begin{tabular}{lrcc}
  \toprule
  Feature & Attribute & Statistical Type & Range/Example \\
  \midrule
  \multirow{2}{*}{Usage} & Usage Level 1& Numerical & [0, 69721] \\
  & Usage Level 2& Numerical  & [0, 190039] \\ \hline
  \multirow{3}{*}{Support} & \# of Interaction & Numerical & [0, 12] \\
  & Maintenance Type & Categorical (4 types) & Scheduled/Unscheduled\\
  & Operation Type & Categorical (7 types) &  Inspection/Repair\\
  \bottomrule
  \vspace{-0.7cm}
  \end{tabular}
  \vspace{-0.2cm}

\end{table}

For example, analysts may consider two types of features: the \textit{usage features} that characterizes customer's service or product usage, and the \textit{support features} that characterize customer's interaction with Customer Engineers (CEs), as depicted in shown in \ref{tab:CCP_attributes}. Real-world datasets with such features are often collected over time and may suffer data quality issues. It is crucial to mine the temporal dependencies from such input sequences, where analysts can find answers to whether or what CE behavior or service (product) usage would help decrease the customer churn rate. Motivated by this use case, we put our focus in this work to the attribution reasoning of which feature and value subspace among the piled-up dimensions has contributed to the classification modeling.

\subsection{RNN and Attention} 

RNN is a family of connectionist models that capture the dynamics of sequences via cycles in the network. 
Recently, RNN-based approaches have demonstrated ground-breaking performance on sequence data analysis such as speech synthesis and time series prediction. 

We use the long short-term memory (LSTM) \cite{Hochreiter1997} recurrent network to capture the features at every time-step conditioned on the previous hidden state. 
Refer to Figure~\ref{fig:training_data}, $x_{t}$ is a vector that encapsulates all the information at one time-step.
We feed LSTM with a sequence of such vectors $X=(x_{1}, \ldots, x_{T})$, T being the length of the sequence, and predict the churning behavior given some hidden states. A hidden state $h_{t}$ is a function $h_{t} = f(W^{x} x_{t}, W^{h} h_{t-1})$. The weight matrices $W^{x}$ and recurrent weight matrix $W^{h}$ are updated through the optimization in the back-propagation process through time. 
The hidden state vector at the final time-step is fed into a binary softmax classifier where it is multiplied by another weight matrix and put through a softmax function that outputs values between 0 and 1, effectively giving us the probabilities of positive and negative customer churn.

The training process with attention mechanism\cite{BahdanauCB14} plugs a vector representation $c$ in between the input layers and the output. The sequence representation takes the form:
\begin{equation}
c = \sum_{t=1}^{T} \alpha_{t}h_{t} \quad \textrm{where} \quad \alpha_{t} = \frac{exp(a(h_{t};W^c))}{\sum_{k=1}^{T} exp(a(h_{t};W^c))}
\end{equation}
where $a(h_{t};W^c)$ is an attention network, and $W^c$ is the parameter set of the attention network. 
We feed the vector $c$ to a softmax classifier to predict the customer churn distribution of the target sequence. We train the attention-enabled model by minimizing the cross-entropy between the predicted distribution and the ground truth. 
A well-trained LSTM model then assigns higher attention scores to the temporal events that are more relevant to the classification task and lower attention scores to the less contributing events. 

During feature encoding, we convert the features in \ref{tab:CCP_attributes} into vector $x_{t}$. We use one-hot vectors $v_{c}$ to represent the categorical attributes, concatenated with the vectors $v_{n}$ that represent the numerical attributes to form $x_{t}$.
After successfully training an LSTM model, the output attention for each time-step of each training instance is associated with the vector $x_{t}$. As a result, the LSTM model assigns the importance for each event in the temporal sequences for the classification task. 
By tracing back to the attribute levels at each event, we achieve the importance of combined feature levels for the whole dataset. We can understand how LSTM is using these features to distinguish the `loyal' from the `leaving' customers by visualizing them in an effective way.

\section{Design Principles}
To visually support the feature attribution reasoning, we consider the following design and development principles.

The design should facilitate the comparison between all features and feature value levels for the temporal events in all training instances.

\textbf{DP2. The design should support the attribution visualization under combined conditions.} Attribute combinations may be more contributive than individual attributes. For example, customer interaction type can be less contributive for the entire customer cohort but can be important for certain types of customers. Therefore, analyzing the combined condition of ’customer type’ and ’customer interaction’ is more meaningful. The design should visualize the contribution of combined conditions in addition to individual variables. 
 
\textbf{DP3. The design should support the visualization of RNN attentions over time.} Customer churn can be the consequence of a series of temporal events. The system design should capture features' contribution from the events with high/low attentions over timesteps. 
On the other hand, from the learned temporal patterns, we expect a better understanding of LSTM's temporal learning features.

\textbf{DP4. The visualization should facilitate the comparison of training instances between different classes, and the comparison of feature attributions.} 
To understand how LSTM uses the features in distinguishing different classes, the visualization should allow users to compare the contributing instances between these classes. The visualization should also support comparing different attributes of data instances.

\textbf{DP5. The visualization and interactions should facilitate the exploration of feature attribution for predictive reasoning.} The system should allow users to explore feature attributions with the flexibility in specifying conditions, such as a particular temporal or attention range. 


\begin{figure*}[htb]
  \centering
  \includegraphics[width=0.9\textwidth]{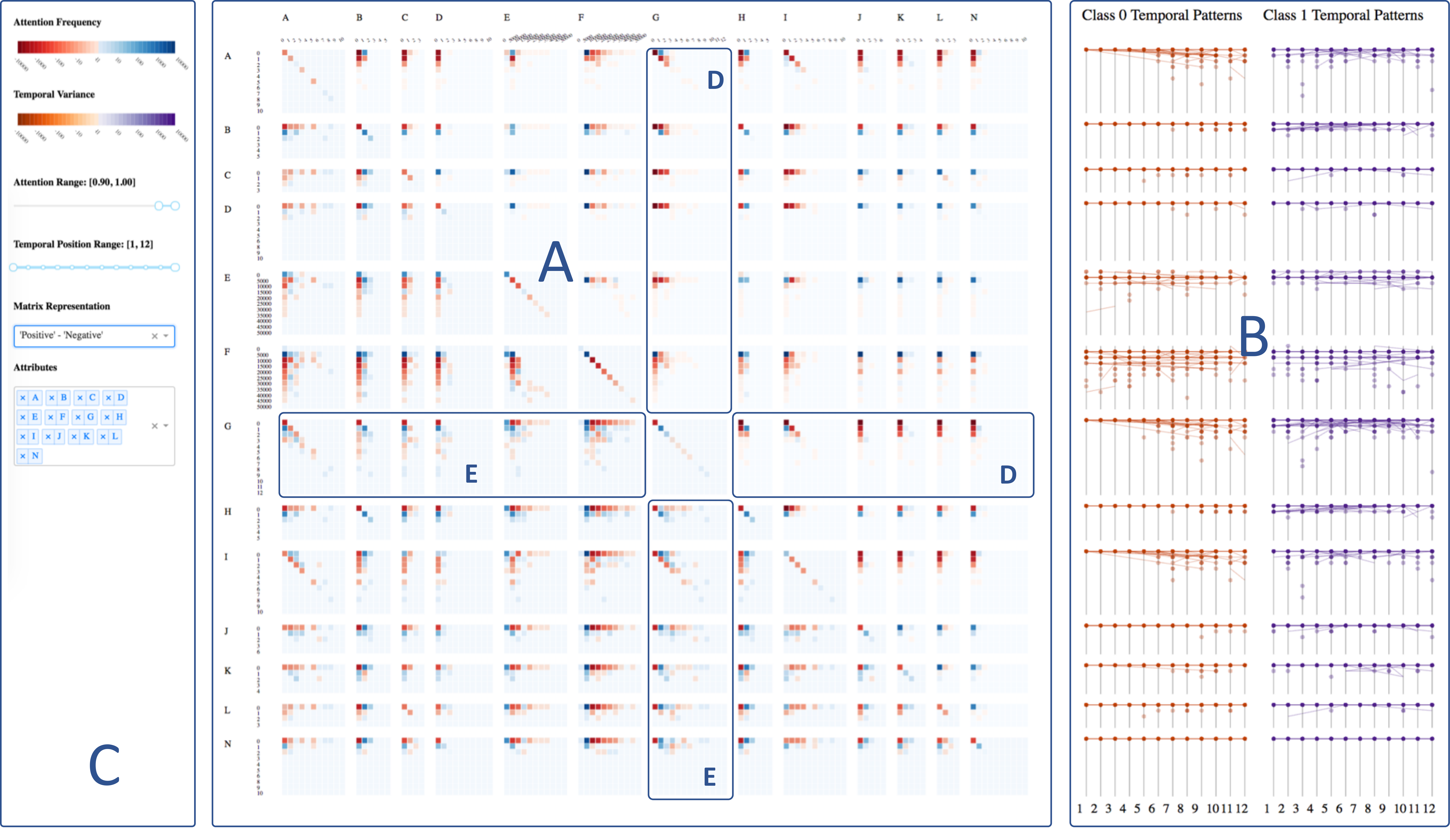}
  \vspace{-0.35cm}
  \caption{\label{fig:system} System overview of AttributionHeatmaps. A is the matrix grid view showing the feature attribution saliences for all training data attributes. The difference from two training classes is shown here. B is the T-partite graph showing the temporal patterns for each feature from two training classes. C is the menu area where users can dynamically analyze feature contributions by viewing and slicing data from different facets. D, E are explained in section 5.}
  \vspace{-0.6cm}
\end{figure*}

\section{Visual Encoding}

We closely work with data analysts and receive feedback during the design iterations. 
Our visual encoding design and the interactions are as follows.

\subsection{Multivariate Feature Visualization}

For DP1 and DP2, we visually encode multivariate data that are assigned with model-specified attention weights.
In an early-stage design, we visualize feature attribution with PCP where axes represent attribute types and coordinates on the axes represent the attribute levels. However, the comparisons between lines of different directions increase the mental burden. We then switch to chaining matrices, where matrices represent the saliency of two combined attribute values. Users then have trouble in deciding matrix chaining orders.

We settle with a concise yet informative design - matrix grid.
As shown in Figure~\ref{fig:system}, area A shows a matrix grid representation of attribute values. Each matrix (excluding diagonal) corresponds to a combination of two attributes. The rows and columns represent attributes' value levels, and each cell reflects a combination of the unique value levels from two attributes. 
To the top and left labels in the matrix grid represent attribute names $[A, B, ..., N]$. Users can further reference the attribute value levels by the labels associated with matrix cells.
The value levels are arranged in the ascending order from left to right and from top to bottom. 
In this paper, we refer matrix $p$\textendash$q$ to the matrix that in column $p$ and row $q$.

\begin{figure}[th]
  \centering
  \vspace{-0.2cm}
  \includegraphics[width=0.95\columnwidth]{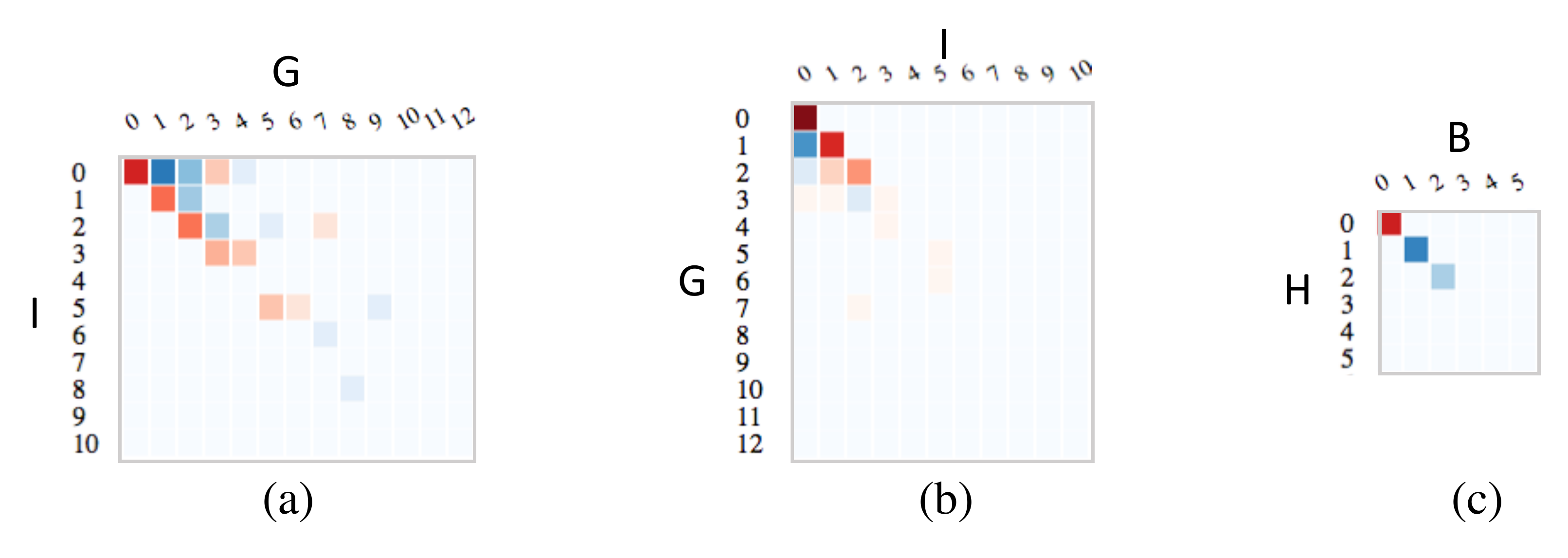}
  \vspace{-0.5cm}
  \caption{\label{fig:pair_matrices} Matrices for combined attributes (a) G-I, (b) I-G and (c) B-H.}
 \end{figure}
 \begin{figure}[th]
  \vspace{-0.5cm}
  \includegraphics[width=\columnwidth]{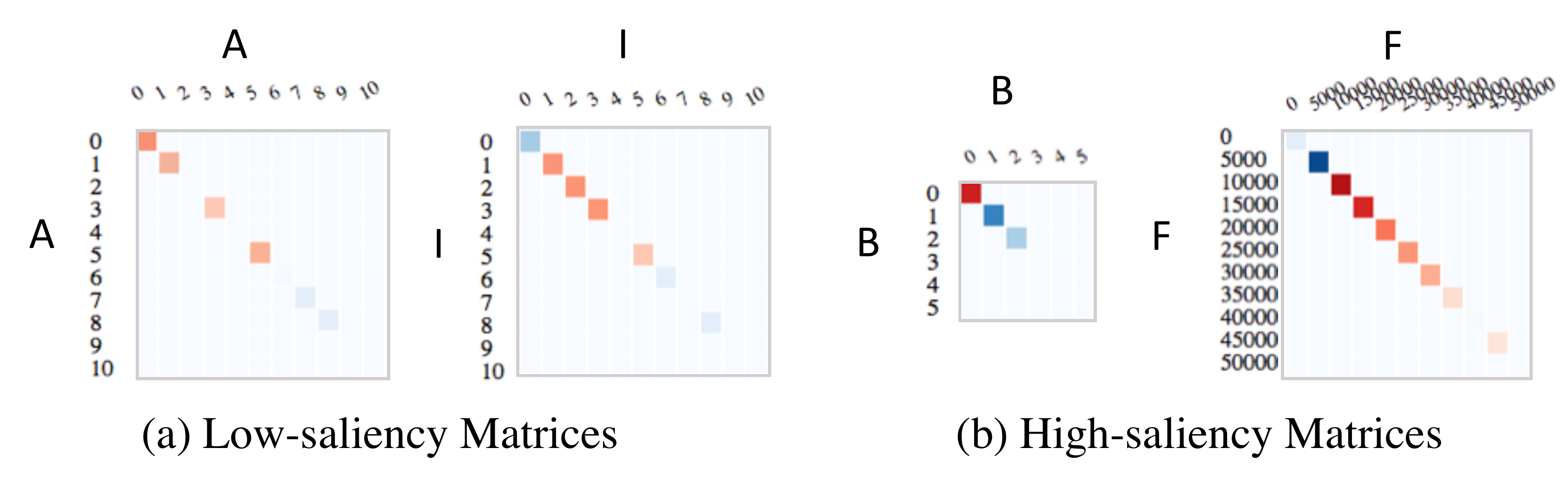}
  \vspace{-0.9cm}
  \caption{\label{fig:individual_matrices} Matrices for individual attributes. (a) attribute A and I exhibit low saliences on the diagonal cells, (b) attribute B and F exhibit high saliences on the diagonal cells.}
  \vspace{-0.8cm}
\end{figure}

The system automatically detects unique value number for categorical attributes on data loading. Color intensity represents the value in a cell, with the upper and lower triangular having different meanings. Cool and warm colors represent positive and negative values, respectively. A user can select the meaning of each cell from `positive class instances,' `negative class instances,' `both classes,' and their difference.

The matrix grid in Figure~\ref{fig:system} shows the difference between positive (blue) and negative (red) instances. High color intensities indicate the instances strongly support their belonging class. 
As in correspondence with DP4, we use explicit encoding \cite{Gleicher2011} to visualize the difference between two classes for feature attribution comparison. 
Users make sense of the highly contributive attributes by locating the matrices of high intensities. The visualization also shows the contribution at a finer granularity: the value level. Users can make sense of contributing values by the color and intensity of the corresponding cells.
The visualization result in area A tends to have more salient values on the top/left areas in many matrices. Because the dataset contains features such as the time of particular activities, for which the number of instances exponentially decays while the number of activities increases. However, the visualization is designed for general purposes, and the trends may differ for different datasets.

The matrices in the lower matrix grid triangle represent the feature attribution saliences. The color intensity represents the absolute value. The logarithmic-scaled colormap is shown on the top of the menu bar. For example in Figure~\ref{fig:pair_matrices} (a), matrix $G$\textendash$I$ represents the feature contribution for two CE activity types $G$ and $I$. The value levels corresponding to each cell represent the number of the activity per month. The saliences concentrate in the upper triangle, which indicates that the number of $I$ is greater than or equal to $G$ for all current data instances. 
The cells on the diagonal exhibit negative contributions as shown in the red cells. 
The intensity decrease from the top left to the bottom right. 
This phenomenon indicates that the instances turn to be predicted by LSTM as ``negative'' when $G$ and $I$ have an equal number of activities in a month, especially for small numbers of activities. The dark red cell on the top left corner indicates that the data instances, where neither $G$ nor $I$ happens in one month, contribute significantly to the negative class.
Also, the cells near the upper diagonal cells, where $G$ values are slightly larger than $I$, show positive contributions in blue colors, which indicates such data instances contribute more to the positive class.

The diagonal cells on the diagonal matrices $P$\textendash$P$ show the contributions of value levels for attribute $p$. As shown in Figure~\ref{fig:individual_matrices}, matrices $B$\textendash$B$ and $F$\textendash$F$ show higher color intensities which indicates a higher contribution to the LSTM classification,
while $A$\textendash$A$ and $I$\textendash$I$ behave conversely. 

The matrices in the upper triangle illustrate the temporal variance of corresponding attributes. 
For cell $(p_{i}, q_{j})$ in matrix $P$\textendash$Q$, we compute the variance over time as $\frac{\sum_{t=T_{range}} (heat^{p_{i}, q_{j}}_{t} - \sum_{t=T_{range}} heat^{p_{i}, q_{j}}_{t}/(T_{range}))^2} {T_{range}}$, reflected by the color intensity. 
Similarly, blue and red colors represent the positive and negative class, respectively. The upper triangular matrices are a reflection of whether the attribute values behave differently over time for all training instances. It exhibits an abstraction of temporal information for higher-level comparisons between attribute combinations. Each matrix works as a button that leads to a finer-level visualization that is introduced in the following sections.
For example, Figure~\ref{fig:pair_matrices} (b) shows high saliences by the red $p_{i} == p_{j}$ cells, which indicates that when the value from $G$ and $I$ equal to each other, instances from the negative class exhibit a higher temporal variance. 

\subsection{T-partite Graph for Temporal Sequences}

In correspondence with DP3, we design a visualization that extends the matrix grid, identifying the difference between training data sequences, and reveals the feature level changes over time. We use the name T-partite graph because the events are partitioned into independent subsets by time.

\subsubsection{Visualization for Individual Attributes}

\begin{figure}[htb]
 \centering 
 \includegraphics[width=\columnwidth]{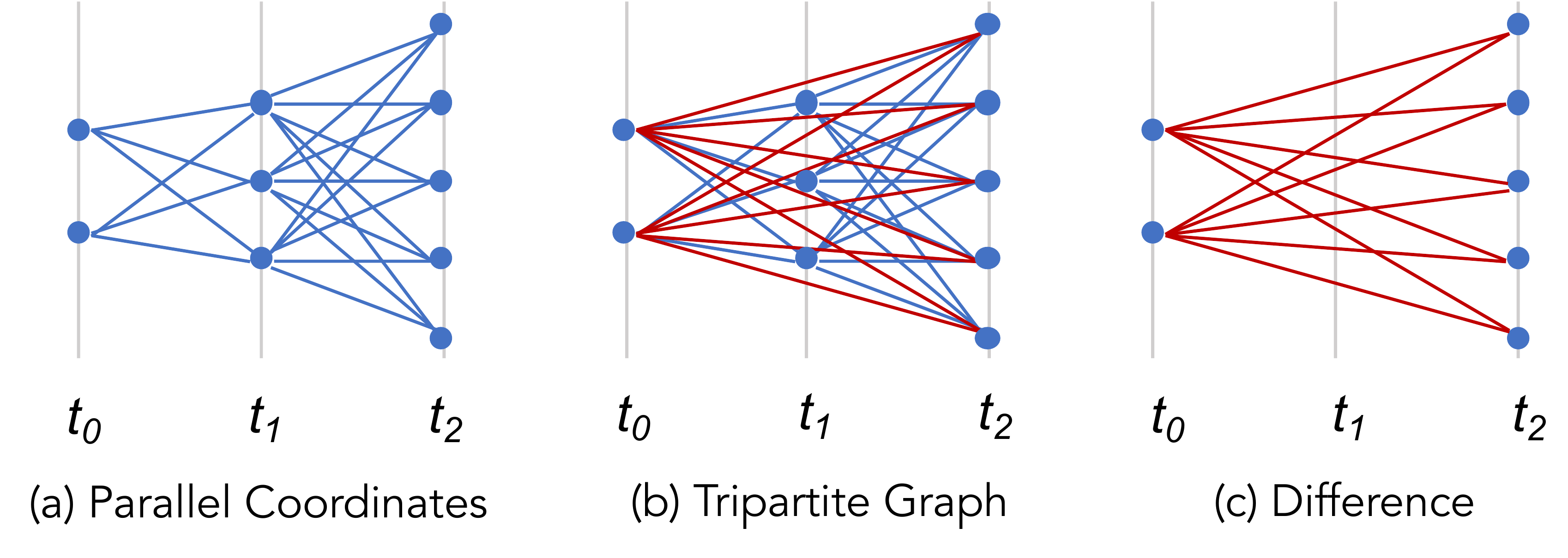}
 \vspace{-0.8cm}
 \caption{The comparison between parallel coordinates and T-partite graphs. Both (a) parallel coordinates and (b) tripartite graph visualize the temporal sequences in time-steps $t_{0}$, $t_{1}$, $t_{2}$, with 2, 3, 5 coordinates on each axes, respectively. (a) visualize all possible event sequences. (b) is a complete tripartite graph that also visualizes all possible event sequences including consecutive temporal sequences and the sequences that across over time-steps. (c) the difference between (a) and (b).}
 \label{fig:PCP_partite}
 \vspace{-0.2cm}
\end{figure}

PCP is proved highly effective by many approaches for visualizing temporal/sequential patterns in DNN\cite{StrobeltGHPR16}.
In these cases, the axes often represent the time-steps, and the events for all sequences are continuous through the time dimension, so the polylines connecting the coordinates on the axes stops at each axis.
In our scenario, because the analysis would involve data filtering with attention weights for temporal events, the filtered event sequences can jump over time-steps. By keeping the advantages of PCP, we visualize the temporal changes in training instances with T-partite graphs, where $T$ equals the maximum number of time-steps.
We illustrate the difference between PCP and T-partite graphs in Figure~\ref{fig:PCP_partite}. 

\begin{figure}[htb]
 \vspace{-0.4cm} 
 \includegraphics[width=\columnwidth]{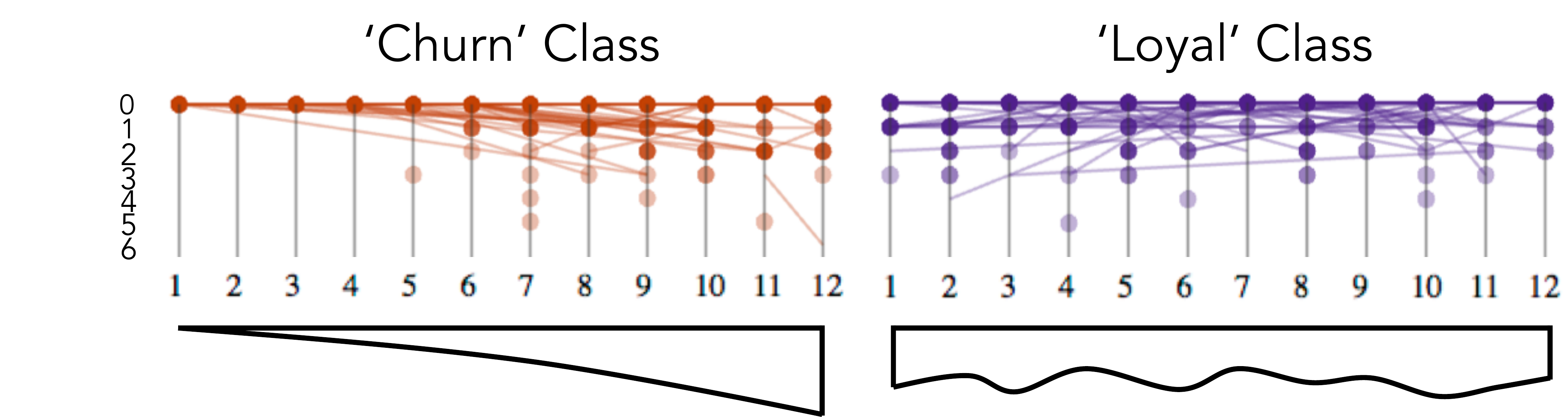}
 \vspace{-0.7cm}
 \caption{The temporal patterns of `the number of maintenance' for the churn and loyal classes, shown in 12-partite graphs. The events from the ``churn'' class show an increasing pattern over time, while the events from the ``loyal'' class are evenly distributed. }
 \label{fig:NOM_temporal}
 
 \vspace{0.1cm}
 
 \includegraphics[width=\columnwidth]{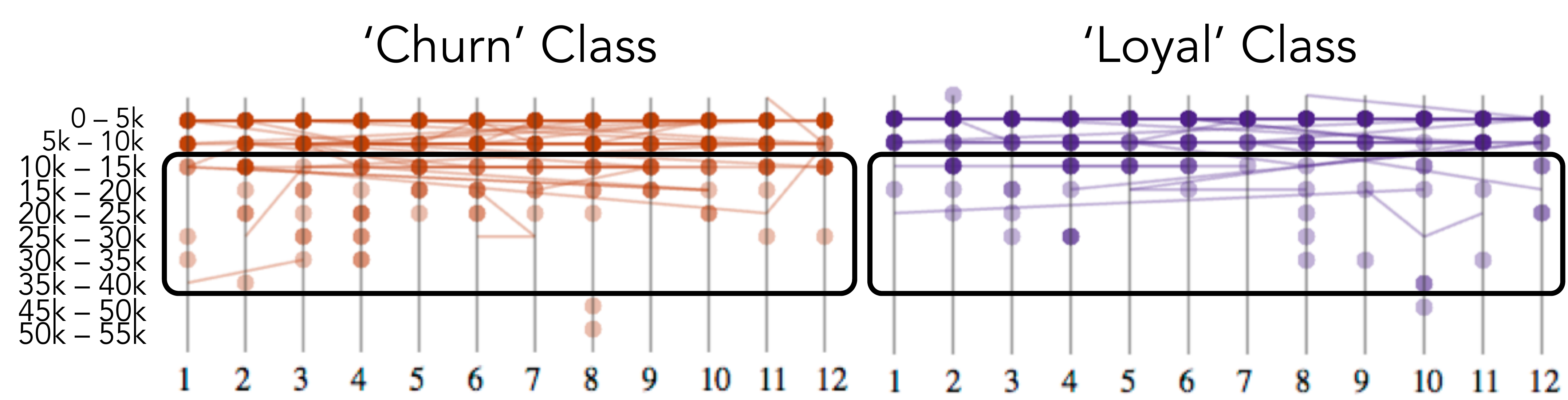}
 \vspace{-0.5cm}
 \caption{The temporal patterns of 'usage level' for the churn and loyal classes, shown in 12-partite graphs. Events from the ``loyal'' class have fewer saliences (especially on the top areas) comparing to the ``churn'' class as shown in the rectangles. }
 \label{fig:UL_temporal}
 \vspace{-0.5cm}
\end{figure}

In Figure~\ref{fig:system} B area, we visualize the  ``churn'' and ``loyal'' temporal sequences in the left and right columns, respectively.
Each row in the T-partite graphs shares the same label as the matrix grid on the left. 
We use the juxtaposition encoding\cite{Gleicher2011}, as shown in Figure~\ref{fig:NOM_temporal} and Figure~\ref{fig:UL_temporal}, to show the difference in temporal pattern from two classes.
Nodes on the axes represent single events from the sequences. We connect events with line segments if there are two or more events selected from a sequence. Purples represent the positive (loyal) class, while oranges represent the negative (churn) class. 
The color representation is consistent with the color coding of the matrix grid view where warm colors represent negatives, and cool colors represent positives.
We encode the number of events with color intensities for nodes and encode the lines with a transparency that is computed based on the maximum number of event frequency. 
Therefore, the significant patterns form saliences due to the highly overlapped lines. 

Figure~\ref{fig:NOM_temporal} and Figure~\ref{fig:UL_temporal} shows the example 12-partite visualization for attribute $number$ $of$ $maintenance$ and $usage$ $level$, respectively. They both exhibit different patterns from two classes, and the patterns distinguish each other for the two attributes. Details are in the explanation under the figures. 

\subsubsection{Visualization for Combined Conditions}

It's challenging to design the visualization in correspondence with DP2 and DP3 due to the complexity of visualizing both multidimensional data and their temporal patterns.

\begin{figure}[t]
 \centering 
 \includegraphics[width=0.6\columnwidth]{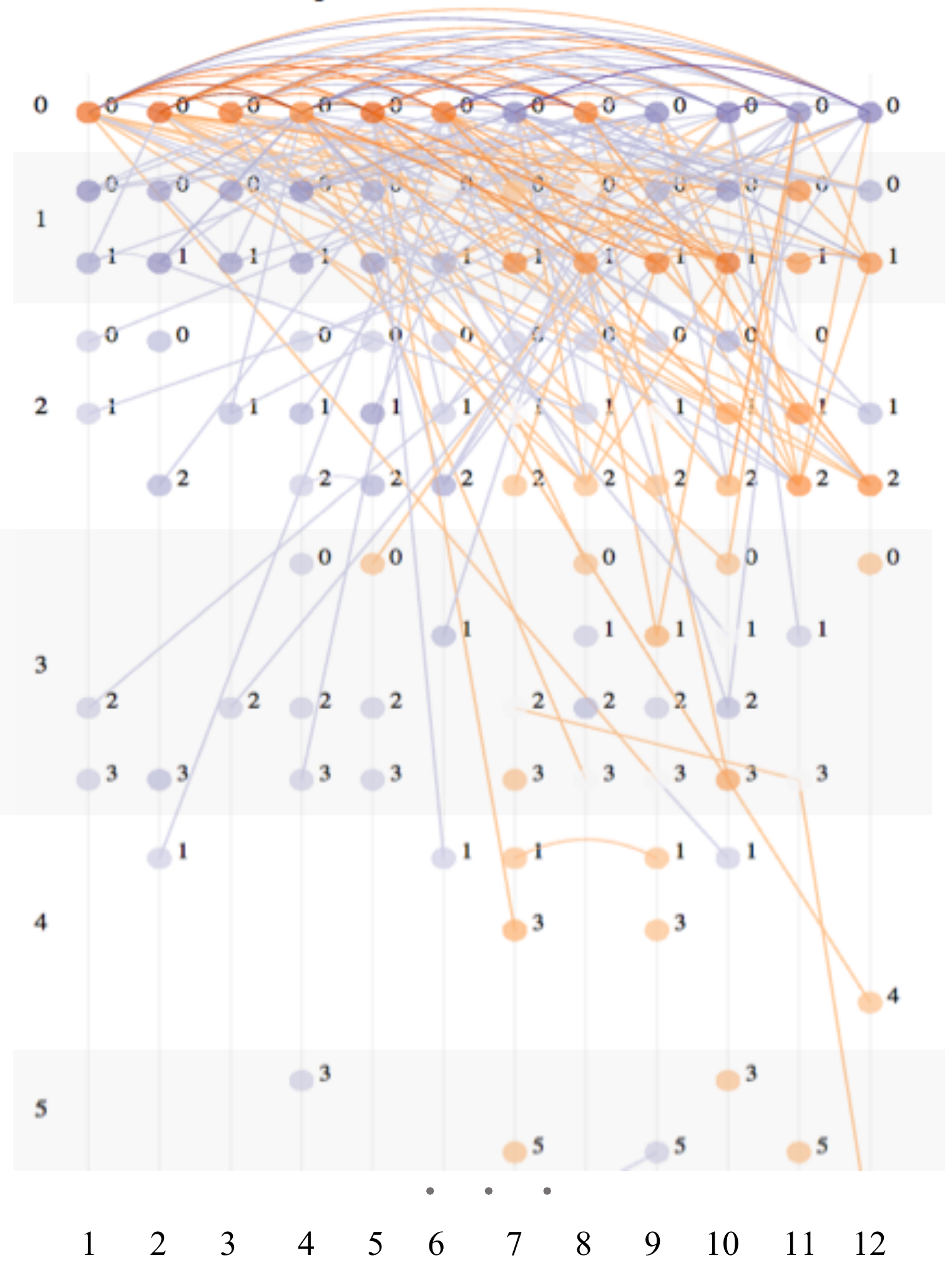}
 \vspace{-0.3cm}
 \caption{Temporal patterns of combined attributes $G$\textendash$A$. }
 \label{fig:temporals}
 \vspace{-0.7cm}
\end{figure}

Figure~\ref{fig:temporals} shows our design for visualizing two attributes $G$ and $A$. In this T-partite graph, axes remain the same meaning - time-steps. 
On each axis, we show the attribute levels in two hierarchies. 
The value levels of the first attribute $G$ are top-down arranged to the left. The number of values is automatically extracted in real-time, and the positions are computed so that the groups use the vertical space efficiently.
The value levels of the second attribute $A$ are top-down listed within the groups of $G$'s levels.
We arrange the vertical positions by calculating the maximum number of values and evenly distributing them within each primary attribute value group along the vertical space. 
Our design also benefits from the mental easiness of referencing the same value levels at the same vertical positions for different time-steps.
However, the edges will overlap each other when the edges connecting nodes with the same value levels. Therefore, we use Bezier curves for such cases so that edges between farther nodes share longer curving edges.
To show the saliency clearly, we sort the edges by their frequency and render the lower-value edges earlier and higher-value edges later.
We use explicit encoding for the comparison between two classes. The number of value levels for combined attributes is greater than the individual attributes. Therefore the comparison in a juxtaposition manner will cause mental burden when viewing back-and-forth between graphs.

The visualization in Figure~\ref{fig:temporals} contains three patterns: I) the lines connecting the nodes on the top left row and the nodes on the lower right half, II) the lines connecting the nodes on the left half area and the nodes on the top row on the right half area, and III) the curves connecting the nodes on the top row. We see the majority of I) are in orange colors and the majority of II) are in purple colors. This phenomenon shows that the decreasing of $G$ values along time has a negative contribution and vise versa.
Within each $G$ value group, we can see that the orange lines mostly connect to large $A$ values. This indicates that small $G$ values with large $A$ values have more negative contributions. 
The orange curves on the right half area in III) shows the highest saliency which indicates that the instances turn to be negative when both $G$ and $A$ are zeros for months 7-12. Also, longer curves have relatively lower intensities, which indicating that long-term events have lower influence comparing to short-term consecutive events. 

\subsection{Feature Contribution Exploration and Reasoning}
In correspondence with DP5, we design the multi-aspect data slicing functionality for the free exploration of feature attribution through the training set.
Our system employs a data-driven architecture where the data controls the flow by offering different data slices chosen by the users. 
We pre-compute the attention values from the LSTM model in the backend and computations triggered by interaction are handled in the front-end to ensure interactivity for the matrix grid and T-partite graphs.
The triggers are in three aspects. 
From the menu area (C in Figure~\ref{fig:system}) users can select data of interest by updating: 

\textbf{The attention range.} 
As introduced in the background, each temporal event from a sequence instance is associated with an attention value that reflects the importance for LSTM to differentiate the characteristics of sequences that lead to customer churn from preserving. 
The analysis benefits from attention-based data slicing in two facets. 
First, comparisons under high and low attention patterns help understand how LSTM make sense of data in the classification, which also helps users solve real-world problems by showing what behavioral attributes assist customer preserving. 
Second, it helps the attribution reasoning under noises. Slight change of filtering conditions may cause significant change due to a blob of highly noisy instances. Users make sense of contribution's changing pattern while sliding the attention range monotonically. Even if there is a sudden change in the process due to noises, the method still can help discover the overall trend.

\textbf{The temporal focus.} As a routine practice, analysts often divide time into stages. In customer churn prediction, data covers over a 12 months span is often explored on a quarterly basis because the attributes are related to CEs for whom customer interactions are organized in such a unit. Also, separating the temporal stages help focus on short-term patterns.

\textbf{The attribute list.} In practice, scientists often train LSTM models with all available attributes when there is no prior knowledge for feature selections. However, not all attributes are contributing to the classification. After generating the visualization and acquiring saliency information, users can remove attributes of low attribution and focus on a subset. In customer churn analysis, the training dataset is initially merged from two separate datasets. For examples, from matrix $C$\textendash$L$ in Figure~\ref{fig:system}, we can observe a high correlation between these two attributes and remove $L$ from the beginning. Users can adjust the sequence of displayed attributes in a certain order for hypothesis verification, or arrange similar attributes for easy locating of common patterns from multiple attributes.

A typical exploration starts from the matrix grid visualization. By default, the matrix grid shows the saliences in contributing features for all attributes. Based on the visualization results, users then explore the data slice of interest with the above-listed interactions. When some attributes of interest are found, they further look into their temporal patterns with the upper triangular matrices and the graphs. 
Due to the complexity of temporal patterns for combined attributes, their visualization requires large canvas space. Clicking the corresponding matrix will trigger the T-partite graph visualization for combined attributes. Clicking the symmetric matrix will exchange the primary and secondary attributes.

\begin{figure*}[tbh]
\vspace{-0.6cm}
 \centering 
 \includegraphics[width=\textwidth]{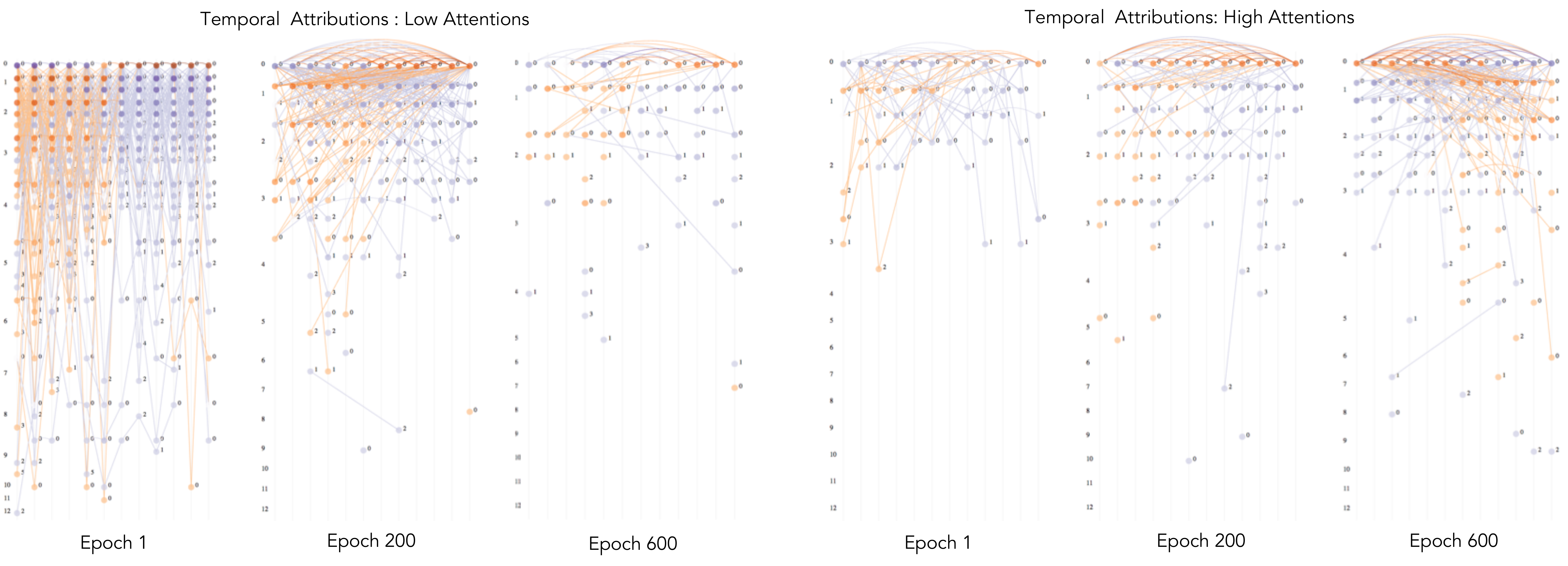}
 \vspace{-0.7cm}
 \caption{LSTM's learning progress: the contribution for combined attributes for high attentions ([0.6, 1.0]) and low attention ([0, 0.2]) at different epochs.}
 \label{fig:epochs}
 \vspace{-0.5cm}
\end{figure*}

\section{Experiment and Evaluation}

The dataset contains 31k events from 2584 customers. 
Each temporal event sequence from a customer is labeled with \textit{churn} or \textit{loyal}. 
Every temporal event contains 14 dimensional features as shown in \ref{tab:CCP_attributes}.
\ref{fig:system} shows the visualization of all attributes. 
The LSTM model was trained using a 9:1 train-test split. The real-world dataset is noisy but we achieved an accuracy of 0.78 around the 600th epochs. 
We normalize the attention vectors to [0, 1] for comparison.

\vspace{0.2cm}
\noindent
\textbf{1) Change Over the Epochs}

Figure~\ref{fig:epochs} illustrates the feature attribution change over the learning process.
The left-most and right-most figures in low and high attention groups represent the first training epoch and the epoch where the model reaches the optimum, respectively. 
The low-attention group shows that almost all sequences are considered unimportant at the beginning, as the graph is nearly complete. The high-attention group shows only a few sequences are considered important at the beginning. While the number of epochs increases, LSTM gradually updates parameters and trims from a large number of sequences for the unimportant sequence patterns and picks up the important ones. 
This explains that the LSTM training process tries to narrow down the conditions that distinguish one class from the other.
Besides, the difference between the positive and negative classes becomes more salient while the number of epochs increases. 

\vspace{0.1cm}
\noindent
\textbf{2) Case Studies}

We continue to work with data scientists and domain experts who have decades of domain experiences as we develop AttributionHeatmap. 
We showcase a few case studies as follows:  

\textbf{Importance in interactively maintaining customer relationship.}
Almost all matrices related to maintenance and operation show red on the top left corner in the matrix grid. 
This indicates that not conducting maintenance and operation greatly contributes to customer churn. 
Domain experts explain that maintenances and operations play an essential role in maintaining a healthy relationship with customers. For some customers that are difficult to find a contact person, it is reasonable that their churn rate is higher.
In addition, more cells are in red tones in the upper triangular matrix grid. This phenomenon indicates that the `churn' class have a larger temporal variance than the `loyal' class. The temporal patterns for `interaction frequency' shown by the T-partite graphs also reveal that the `churn' class tend to have higher frequencies in the first time-step and the frequencies decrease in later time-steps. Meanwhile, the `loyal' class have more evenly distributed frequencies among all time-steps. The visualization indicates that it is easier to retain a customer from churning if interacting with the customer at a stable frequency. 
The visualization also reveals churn-contributing patterns where interactions are highly infrequent in the first-half and increase greatly in the second-half temporal period.

\begin{figure}[th]
 \centering 
 \vspace{-0.5cm}
 \includegraphics[width=\columnwidth]{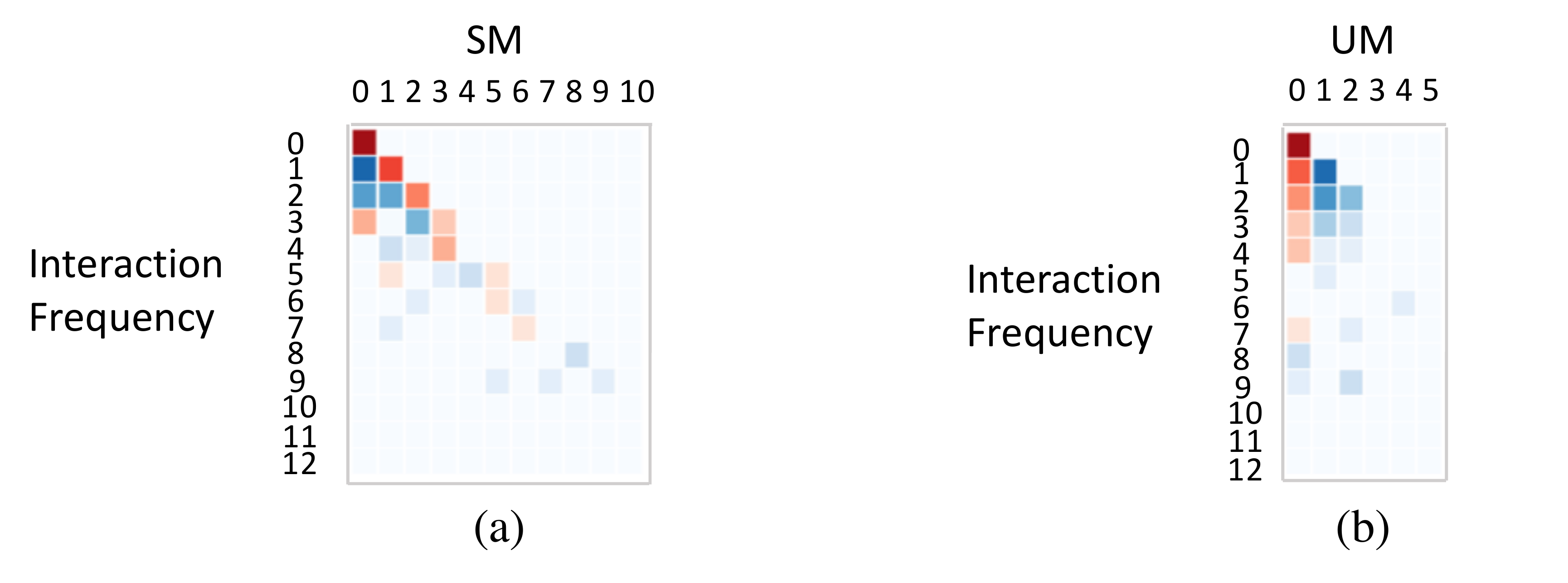}
 \vspace{-0.9cm}
 \caption{The AttributionHeatmaps for (a) SM-IF (b) UM-IF. }
 \label{fig:actCnt-UM}
 \vspace{-0.2cm}
\end{figure}

\textbf{Customer Interaction: Scheduled or Unscheduled?}
Maintenance can be Scheduled Maintenance (SM) or Unscheduled Maintenance (UM). 
UMs are for unexpected issues, and SMs are routine and periodic. Interaction frequency (IF) covers all maintenance types including SM and UM.
Common sense is that UM has negative contributions because UM indicates some issues in the products. However, AttributionHeatmap interprets contrarily from several perspectives. Figure~\ref{fig:actCnt-UM} (a) and (b) show the heatmap for SM-IF and UM-IF. In (a), the diagonal red cells show that interacting with customers only in SMs contributes to customer churn.
The blue cells under the red line show that other interaction types contribute otherwise. 
In contrast, the first red column in (b) shows that not conducting UM has negative impacts. 
Besides, the dark blue 1-1 and 1-2 cells indicate that having once UM per month have a greater positive impact. 
These two matrices together reveal that UM has a higher contribution in customer retention and SM highly contributes to customer churn. 
Explanations are when customer issues are not solved at the first interaction, the next interaction becomes SM. 
Thus one UM in a month, in a high chance, indicates that CEs solved the problems from customers at one interaction. Therefore, the observation verifies that quick problem solving is important in customer retention.
The T-partite graphs of IF (primary) and UM (secondary) show bottom-left-to-top-right purple edges and top-left-to-bottom-right orange edges, which means more interactions in the first half year and fewer interactions in the second half year contributes to the customer retention. The visualization results also show that one-time UM in the second year (especially month 10) have negative contributions, which provide hints to guide CE interactions.

\vspace{0.1cm}
\noindent
\textbf{3) Accuracy Improvement}

From the case studies, we found the usage features E, F, and the support features B, H, J have strong attribution, where the usage features are more important than the support features. 
The found important features guide data engineers to source more fine-grained data for the support features. 
During the second experimental iteration, the customer churn prediction accuracy is improved by 0.05 after including such data. 
Although the improvement is small, scientists give highly positive feedbacks since they have tried to train the model with many attributes but the model fails in getting a better accuracy. This proves our visual reasoning approach helps reveal feature attribution and suggest fine-grained factors for the improvement of prediction/classification accuracy.

\section{Discussion}
We discuss the extensibility and limitation in this section.

\textbf{Generality}
The use of AttributionHeatmap is not limited to the feature attribution for customer churn prediction. The data-driven approach we adopt does not depend on any application-specific condition and can be generalized to more applications. Our approach also handles different statistical data types automatically.
Besides, the T-partite graph depends on the matrix grid, but not contrariwise. The matrix grid view alone works as a visualization view for feature attribution without digging into the temporal changes. Therefore, the matrix grid can be potentially extended to the feature attribution analysis for Convolutional Neural Networks. 

\textbf{Scalability}
AttributionHeatmap uses WebGL-based rendering and therefore supports big data analysis.
We demonstrate in this paper that the dataset contains 14 attributes, each of which has 9 value levels on average. And each data instance contains 12 temporal events. The ability to handle such a complex dataset demonstrate the scalability of our system.
Also, the matrix grid view expands while the number of attributes and the corresponding value levels increase, but the UI lets users trim and rearrange rows and columns during the exploration.

\textbf{Limitations}
AttributionHeatmap works with fine-trained RNN models. Our approach does not apply for the datasets that fail in learning a convergent model.
Besides, the higher the prediction accuracy is, the better the conclusion can be made. Without an accurate prediction model, the visualization result would suffer from lacking reasonable evidence.
Moreover, AttributionHeatmap works best for balanced datasets because it compares the data instances' contributions from different classes. For imbalanced datasets, conclusions can be made based on oversampled or undersampled data.

\section{Conclusion}
\noindent
In this paper, we present an effective solution for value-level feature attribution analysis with labeled multivariate temporal sequences. We implement a functional tool \textit{AttributionHeatmap} with an easy-to-interpret design. AttributionHeatmap facilitates the understanding of contributing features and their temporal patterns for predictions. We deploy and test AttributionHeatmap with a churn prediction dataset collected over 12 months from an international corporation. The experimental results help to understand the learning progress of attention mechanism in RNNs. The case study results demonstrate that the approach can help domain experts reason feature attribution and give suggestion to achieve their goals.

\section*{Acknowledgment}
\addcontentsline{toc}{section}{Acknowledgment}
This work was supported in part by the U.S. National Science Foundation with grant   IIS-1741536.

\bibliographystyle{IEEEtran}
\bibliography{IEEEabrv,attention}

\end{document}